\documentclass{llncs}

\usepackage[T1]{fontenc}

\usepackage{graphicx}
\usepackage{subfigure}
\usepackage{subcaption}
\usepackage{algorithmic}
\usepackage{booktabs}
\usepackage{multirow}
\usepackage{booktabs} 
\usepackage{hyperref}
\usepackage{multirow}
\usepackage{wrapfig}

\usepackage{amsmath}
\usepackage{amssymb}
\usepackage{mathtools}
\usepackage{algorithm}
\usepackage{algorithmic}
\usepackage{tikz,pgfplots,diagbox}

\usetikzlibrary{positioning}
\usetikzlibrary{patterns}

\newcommand\name{\textsc{DiverseDistill}}
\definecolor{sunny}{rgb}{0., 0., 0.}
\AtBeginDocument{%
  }

\begin{document}

\title{Wisdom of Committee: Diverse Distillation from Large Foundation Models and Domain Experts\thanks{Accepted at the 1st Workshop on 
Resource-Efficient Learning and Knowledge Discovery 
(RelKD), KDD 2026.}}

\author{
Zichang Liu\inst{1}
\and Qingyun Liu\inst{2}
\and Yuening Li\inst{3}
\and Liang Liu\inst{3}
\and Anshumali Shrivastava\inst{1}
\and Shuchao Bi\inst{3}
\and Lichan Hong\inst{2}
\and Ed H. Chi\inst{2}
\and Zhe Zhao\inst{2}\thanks{Now at University of California, Davis. Email: zao@ucdavis.edu}
}

\institute{
Rice University
\email{\{zichangliu, anshumali\}@rice.edu}
\and Google DeepMind
\email{\{qyl, lichan, edchi\}@google.com}
\and Google Inc
\email{\{yueningl, liangliu, shuchaobi\}@google.com}
}

\maketitle
\thispagestyle{plain}

\pagestyle{plain}

\begin{abstract}
    
  Knowledge distillation from foundation models to compact domain
models is challenging due to substantial gaps in capacity,
architecture, and modality. For example, in our experiments, distilling from a 76M-parameter language model to a 2M-parameter recommender closes less than 40\% of the performance gap
between the undistilled student and the teacher. We show that introducing domain-specific experts---which share
the student's architectural characteristics---alongside the
foundation model as a diverse teacher committee significantly
improves transfer. However, standard
multi-teacher methods fail to exploit this diversity: naively
combining heterogeneous teachers can degrade performance below
single-teacher distillation. To address this, we propose
\name{}, an interactive distillation framework that employs a
learnable Question--Answer mechanism to generate
teacher-conditioned queries and align heterogeneous teacher
outputs into the student's representation space. Unlike methods requiring gradient-based co-optimization or architectural modification of teachers, \name{} operates with frozen teachers using only forward-pass inference through their intermediate layers: no parameter updates, no co-training, and no architectural surgery. A dynamic teacher importance mechanism further
reduces training cost by filtering low-relevance teachers per
sample ({\em e.g.} $\sim$30\% fewer forward passes with no quality loss for recommendation tasks),
while the entire Distillation Module is discarded after training,
adding zero inference overhead. Evaluations on recommendation
(38$\times$ compression) and vision (3.6$\times$ compression)
tasks demonstrate that \name{} recovers 73--114\% of the
teacher--student performance gap, consistently outperforming
all single- and multi-teacher baselines.

\end{abstract}

\section{Introduction}

The rise of foundation models has revolutionized machine learning, showing remarkable performance and generalization across diverse tasks~\cite{radford2021learning,dosovitskiy2021image,brown2020language,devlin2019bert,ramesh2021zeroshot}. Trained on massive, web-scale data, these models can be adapted to new applications through fine-tuning~\cite{hu2021lora,devlin2019bert} or leveraging their few-shot or in-context learning ability~\cite{brown2020language,bai2023sequential}.

However, specialized domain models also have distinct advantages. Optimized for specific tasks like financial analysis or personalized recommendation~\cite{tang2018ranking,kang2021topology,tsantekidis2021diversity}, these models are highly efficient and excel at memorizing task-specific data. For instance, while large language models (LLM) demonstrate strong generalization for recommendation tasks~\cite{kang2023llms}, traditional embedding-based recommender systems (RecSys) are the production standard, optimized with domain expertise for performance on specific distributions~\cite{naumov2019deep,liu2023multitask}.

To harness the strengths of both, a natural progression is to transfer knowledge from powerful foundation models to more efficient specialized domain models~\cite{xu2025slmrec,liang2025fm}. This process, a form of knowledge distillation, faces unique challenges due to the fundamental differences between the two model types. Some main challenges include: substantial gaps in model capacity which hurts distillation~\cite{cho2019efficacy,mirzadeh2020improved}, differing architectures and input formats~\cite{yun2019transformers,edelman2022inductive} ({\em e.g.}, Transformers vs. embedding-based models), misaligned training distributions~\cite{huo2024c2kd,wei2024promptmm}, and the use of different features ({\em e.g.}, text vs. categorical features) that creates issues with privileged information~\cite{yang2022toward,shi2024cpfd}.

To address the challenges of directly distilling knowledge from a
foundation model to a specialized student, we introduce
domain-specific models as complementary teachers. This forms a
diverse teacher committee that can bridge the aforementioned gaps:
domain teachers, with model capacity and architecture closer to the
student's, serve as intermediate knowledge carriers that facilitate
smoother transfer~\cite{park2022learning,mirzadeh2020improved},
while foundation models contribute broader generalization and
cross-modal understanding. Analogous to a Ph.D. committee with
varied expertise, this diversity allows the student to draw on
complementary knowledge sources rather than relying on any single
teacher.

\begin{figure}[t]
  \centering
      \includegraphics[width=0.8\textwidth]{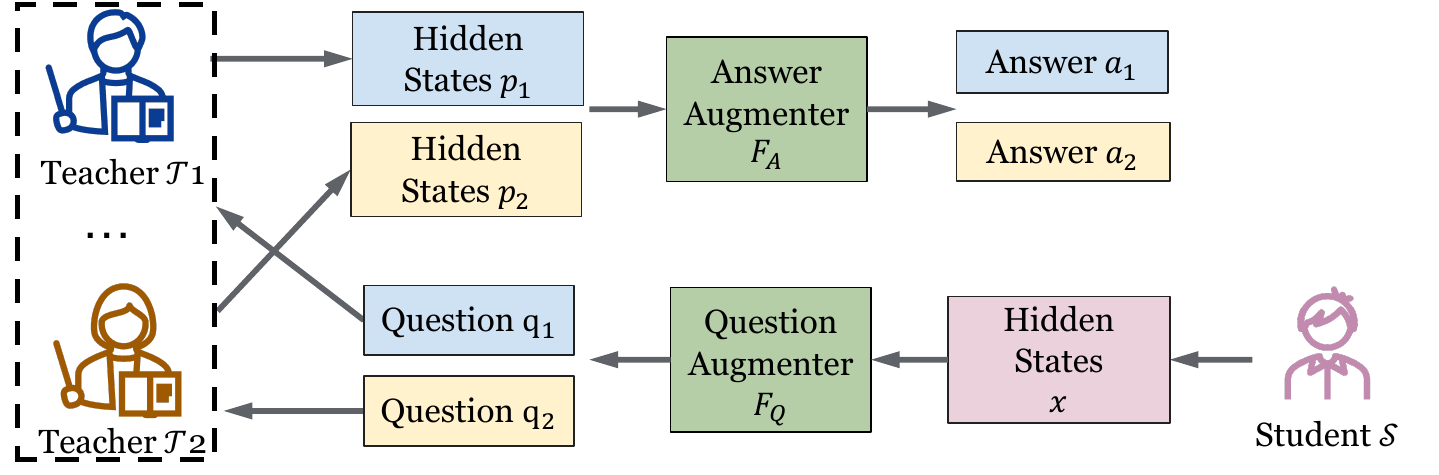}
  \caption{Overview of \name{}. Teachers (left) form a diverse
committee. The Distillation Module (green) comprises a Question
Augmenter $F_Q$ and an Answer Augmenter $F_A$. $F_Q$ generates
teacher-specific queries from the student's representation,
conditioned on learnable teacher embeddings. $F_A$ projects each
teacher's response into the student's hidden space, producing
aligned soft targets for the distillation loss.}
  \label{fig:workflow} 
\end{figure}

However, this diversity introduces a new challenge: standard
multi-teacher distillation methods frequently \emph{fail} when
teachers differ substantially in architecture, capacity, and input
modality---particularly when foundation models are involved.
For example, simple soft-label averaging across heterogeneous
teachers may degrade the student below single-teacher performance
on multiple benchmarks (Sec.~\ref{sec:experiment_result}).
This indicates that realizing the benefits of teacher diversity
requires an explicit mechanism to selectively extract and align
each teacher's unique expertise.

To this end, we propose \name{}, an interactive distillation
framework that enables teacher-conditioned query generation to selectively extract relevant knowledge and generate tailored queries to selectively extract customized
knowledge, as shown in Fig.~\ref{fig:workflow}. A key design property of \name{} is that it operates with
frozen teachers, requiring only forward-pass inference through their
intermediate layers---no gradient propagation into teacher parameters,
no architectural modifications, and no co-training. This is
particularly valuable when teacher models are independently
pre-trained, maintained by separate teams, or shared across multiple
applications. More broadly, \name{} adapts to arbitrary combinations
of teachers and students regardless of differences in architecture,
capacity, data distribution, or input features, relaxing the
restrictive assumption in existing multi-teacher methods that
teachers must share similar characteristics or be co-optimized
with the student~\cite{10.1145/3097983.3098135,ding2022skdbert,NEURIPS2020_91c77393}.

Additionally, \name{} provides a mechanism to reduce computational
overhead during distillation. Since foundation models are generally
more resource-intensive, our framework dynamically computes the
relevance of each teacher for a given data sample via learned
importance scores. Teachers with low relevance are skipped,
avoiding unnecessary forward passes---an approach we show can
substantially reduce training cost while preserving model quality
(Sec.~\ref{sec:distillation_design}). Moreover, all auxiliary distillation components are used
only during training and entirely discarded afterward: the
deployed student model is identical in size and latency to an
undistilled model, incurring zero additional inference cost.

In summary, we argue that forming a diverse teacher committee
enables more effective knowledge transfer from foundation models to
specialized application models than single-teacher distillation.
To harness such committees, we propose an interactive
question-and-answer distillation framework with learnable parameters
to model each teacher's expertise. Our main contributions are:

\begin{itemize}

\item We identify a fundamental challenge in distilling from
foundation models to specific domains: direct distillation is often ineffective due to
capacity and modality gaps, yet naively combining foundation
models with domain experts via standard multi-teacher methods
can \emph{degrade} student performance---making diversity a
liability rather than an asset. To address this, we propose
constructing a diverse teacher committee and introduce a
mechanism to unlock its potential
(Sec.~\ref{sec:whole_design}).

\item We develop \name{}, an interactive distillation framework
that resolves the above challenge through a learnable
Question--Answer mechanism. The framework aligns heterogeneous
teacher representations by generating teacher-specific queries
and projecting diverse responses into a shared student space.
It operates with strictly frozen teachers---requiring only
forward-pass access, no co-training, and no architectural
modifications---enabling distillation across arbitrary
teacher combinations ({\em e.g.}, LLM + RecSys, CLIP + ViT)
(Sec.~\ref{sec:distillation_design},~\ref{sec:training_distillation}).

\item We demonstrate that \name{} includes a dynamic teacher
importance mechanism that filters low-relevance teacher computations
per sample. On recommendation tasks, this reduces teacher forward
passes by $\sim$30\% with no performance
loss. After training, all auxiliary distillation components are entirely
discarded, achieving model compression ratios of up to
38$\times$ with zero additional inference cost (Sec.~\ref{sec:end2end},~\ref{sec:exp_ablation}).

\item We evaluate \name{} on recommendation and vision tasks
with heterogeneous committees. \name{} consistently outperforms
all single- and multi-teacher baselines, recovering up to
114\% of the teacher--student gap---surpassing the best
individual teacher---while standard methods fail under the
same diversity conditions
(Sec.~\ref{sec:experiment_result}).

\end{itemize}

The datasets utilized in this work are publicly available~\cite{10.1145/2827872,Krizhevsky2009LearningML,russakovsky2015imagenet}. While the model architectures rely on standard components, the training pipeline and distillation framework are implemented using proprietary internal libraries and distributed computing infrastructure, preventing open-source release. However, we have provided detailed algorithmic descriptions and hyper parameter settings to ensure reproducibility.

\section{Related Work}


\noindent \textbf{Knowledge Distillation (KD).} KD is a promising, hardware-friendly direction for reducing the serving costs of deep neural networks~\cite{hinton2015distilling}. However, significant differences between the teacher and student models can hinder knowledge transfer. For example, a large capacity gap between the two can degrade student performance, as smaller teachers are sometimes found to yield better transferable representations~\cite{park2022learning,mirzadeh2020improved}. Another challenge is from the task difference between a pre-trained teacher and the student's downstream task~\cite{zhao2023talking}. In contrast to existing cross-distillation methods that interweave teacher-student hidden layers~\cite{bai2020crosskd,wang2024crosskd}, our work specifically targets knowledge transfer from foundation models to specialized applications. Our methodology introduces a distillation module that enables the student to selectively learn from the teacher, rather than mimicking its entire performance.

\medskip
\noindent \textbf{Foundation Model Distillation.}
Distilling knowledge from foundation
models~\cite{devlin2019bert,brown2020language,ramesh2021zeroshot,radford2021learning}
into compact students has attracted significant
interest~\cite{agarwal2023gkd,hsieh2023distilling}. Early work
focused on compressing within the same model family where architectural
compatibility simplifies alignment, {\em e.g.},
DistilBERT~\cite{sanh2020distilbert},
TinyViT~\cite{wu2022tinyvit}, or text-to-text
distillation~\cite{hsieh2023distilling}. Cross-architecture extensions
exist but remain restricted to specific pairings such as
Transformer--CNN~\cite{liu2022cross,liu2024transkd,ni2024crossarchi}
or autoregressive LMs~\cite{agarwal2023gkd}. In recommendation,
SLMRec~\cite{xu2025slmrec} distills LLMs into smaller language
models, while ~\cite{liang2025fm} leverage external
foundation models as auxiliary signals; both, however, assume the
student belongs to the same model family or require teacher
co-adaptation for alignment.

More recently, distillation from multimodal foundation models~\cite{wu2021teacher} has gained momentum, yet most methods impose assumptions that limit
practical applicability~\cite{sun2023dimefm}. In recommendation, $\text{C}^2\text{KD}$~\cite{huo2024c2kd}
requires partial teacher fine-tuning via bidirectional distillation,
and PromptMM~\cite{wei2024promptmm} injects learnable prompts into
the teacher's internal layers. In vision,
UNIC~\cite{sariyildiz2024unic} distills from multiple foundation
teachers (DINOv2, CLIP, etc.) into a unified student, but assumes
all teachers and the student share a compatible vision-encoder
architecture. In contrast, \name{} targets a more general setting
where teachers and students share minimal characteristics ({\em e.g.}, an
LLM teacher vs.\ a RecSys MLP student), treating all teachers as
strictly frozen experts---requiring only forward-pass access without
internal modifications or co-training.

\medskip
\noindent\textbf{Multi-Teacher and Heterogeneous Distillation.}
Multi-teacher distillation improves student performance under the
philosophy that ``two heads are better than one.'' Traditional
methods combine teacher knowledge by averaging or selecting soft
targets, logits, or intermediate representations~\cite{10.1145/3097983.3098135,park2019feed,asif2020ensemble},
while later work introduced dynamic weighting based on task
loss or prediction consistency~\cite{NEURIPS2020_91c77393,Liu_2020}, student
feedback~\cite{yuan2020reinforced,ding2023autoskdbert}, or probe-set for task
similarity~\cite{borup2023distilling}. Recent methods further
address heterogeneous teacher architectures:
MTKD~\cite{jiang2024mtkd} integrates CNN and ViT teachers for
super-resolution, DCHKT~\cite{zhou2025dynamic} uses dynamic
collaborative learning for long-tailed recognition,
~\cite{lin2025perspective} propose
perspective-aware teaching for heterogeneous distillation, and
~\cite{li2025fuse} fuse multi-teacher knowledge before
transferring it to the student. However, these methods face two
fundamental limitations in our setting:


\emph{(1)~Incompatible output spaces.}
Most multi-teacher methods---including dynamic weighting
approaches~\cite{NEURIPS2020_91c77393,yuan2020reinforced,jiang2024mtkd}
and fusion-based
methods~\cite{li2025fuse,lin2025perspective}---assume
teachers share compatible output spaces ({\em e.g.}, identical label
sets) or permit access to aligned internal features. In a
diverse committee ({\em e.g.}, LLM + RecSys), teachers produce
fundamentally incompatible outputs (vocabulary logits vs.\
ranking scores) that standard fusion cannot resolve.

\emph{(2)~Unavailable training signals.}
Heterogeneous approaches such as
HetComp~\cite{10.1145/3543507.3583209},
MT-BERT~\cite{wu2021teacher}, DCHKT~\cite{zhou2025dynamic},
PromptMM~\cite{wei2024promptmm}, and
UNIC~\cite{sariyildiz2024unic} rely on accessing training
trajectories, co-optimization, prompt injection, or shared
architectural backbones. These requirements are impractical
for foundation models, which are typically deployed as frozen
black boxes whose internals and gradients are unavailable.

\name{} is designed to simultaneously handle \emph{cross-architecture} teachers (MLP,
Transformer, CLIP), \emph{cross-modality} knowledge (text, vision,
tabular), \emph{strictly frozen} teachers (no co-training or
internal access), and \emph{dynamic per-sample importance}
weighting: addressing both limitations above through its learnable
Distillation Module.

\section{Method: Distilling from Diverse Teachers}
\label{sec:method}

\begin{figure}[t]
    \centering
    \subfigure[Question Augmenter $F_Q$]{
    \includegraphics[width=0.53\textwidth]{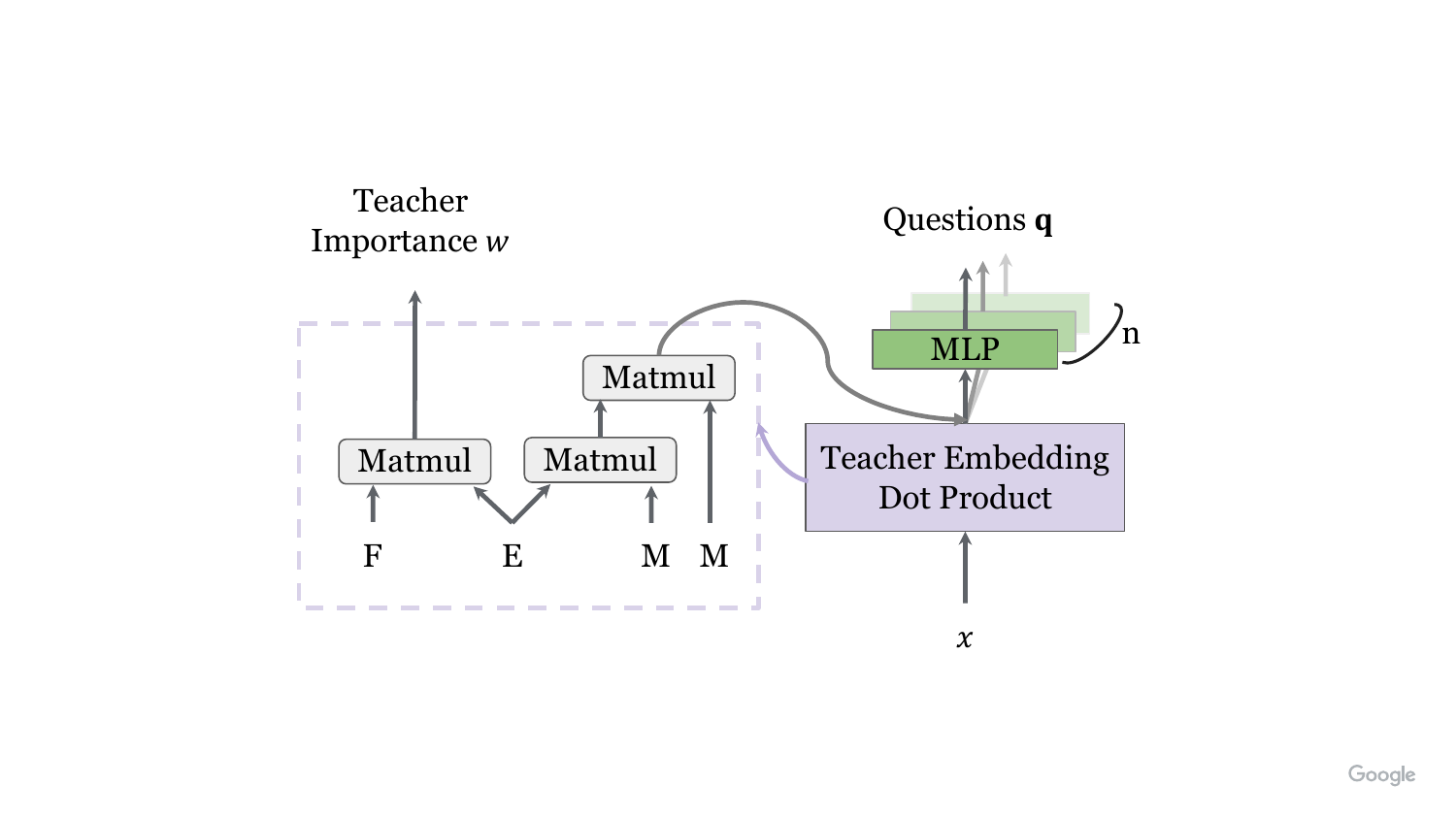}
    \label{fig:question_generator_design}
    }
    \subfigure[Answer Augmenter $F_A$]{
    \includegraphics[width=0.32\textwidth]{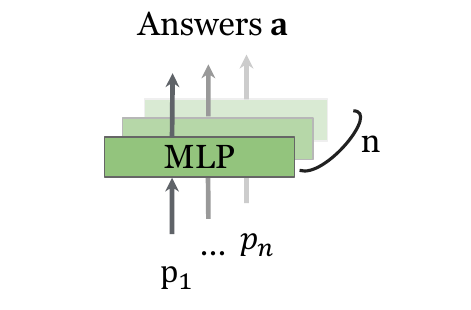}
    \label{fig:answer_generator_design}
    }
    \caption{Details of the Distillation Module.
(a)~$F_Q$ takes the student's representation $\mathbf{x}$ and
produces teacher-specific questions $\mathbf{q}$ and importance
weights $\mathbf{w}$. $E$: learnable teacher embeddings,
$M$: teacher-conditioned question projection,
$F$: per-sample importance scoring.
(b)~$F_A$ maps each teacher's hidden states $p_i$ into the
student's space, yielding aligned answers $a_i$ as soft targets.}
    \label{fig:distillation_module}
\end{figure}

\name{} is a distillation framework that transfers knowledge from
a diverse committee of frozen teachers into a compact student model.
Unlike standard distillation, where teachers unilaterally provide
soft labels via logits or intermediate
features~\cite{mirzadeh2020improved}, \name{} introduces a learnable
\emph{Distillation Module} (Fig.~\ref{fig:workflow}) that enables
teacher-specific communication: it generates tailored queries
conditioned on each teacher's learned characteristics and aligns
their heterogeneous responses into the student's representational
space. The key insight is that
formulating queries in the ``language'' of each teacher's domain
leads to more effective knowledge transfer than treating all teachers
uniformly. Unlike adaptive distillation methods that require
co-training or updating
teachers~\cite{szatkowski2024adapt,zhou2022bert}, all teachers in
\name{} remain strictly frozen, making it applicable to settings
where teachers are pre-deployed or accessed via inference-only
interfaces.

We first describe the high-level distillation process
(Sec.~\ref{sec:whole_design}), then detail the architecture of the
Distillation Module (Sec.~\ref{sec:distillation_design}), discuss
a regularization mechanism critical for training it
(Sec.~\ref{sec:training_distillation}), and finally present the
complete end-to-end training procedure
(Sec.~\ref{sec:end2end}).

\subsection{Distillation Process}
\label{sec:whole_design}

Given a set of pre-trained, frozen teachers $\{T_1, \ldots, T_n\}$,
\name{} jointly learns the student model $S$ alongside the
Distillation Module, which comprises a Question Augmenter $F_Q$
(Fig.~\ref{fig:question_generator_design}) and an Answer Augmenter
$F_A$ (Fig.~\ref{fig:answer_generator_design}). Each teacher $T_i$
may differ in architecture, hidden dimension, and even output space;
$F_Q$ and $F_A$ bridge these mismatches through per-teacher
projection layers (detailed in
Sec.~\ref{sec:distillation_design}). The module operates in three
stages for each data sample $d$:

\begin{enumerate}
    \item \textbf{Generate questions ($F_Q$).}
    A representation $\mathbf{x}$ is extracted from the student $S$
    ({\em e.g.}, concatenated input embeddings or a pooled encoder output)
    and fed into $F_Q$, which produces (i)~$n$ teacher-specific
    questions $\mathbf{q} = \{q_1, \ldots, q_n\}$, each projected
    into $T_i$'s native input dimension, and (ii)~per-teacher
    importance weights $\mathbf{w} = \{w_1, \ldots, w_n\}$. Both
    are conditioned on learnable teacher embeddings that capture
    each teacher's characteristics.

    \item \textbf{Collect answers (frozen teachers).}
    Each question $q_i$ is injected into teacher $T_i$ at an early
    processing stage. $T_i$ performs a forward pass from that point
    and returns its last-layer representation $p_i$.

    \item \textbf{Align answers ($F_A$).}
    $F_A$ maps each $p_i$ from $T_i$'s native space into the
    student's hidden dimension via per-teacher projection layers,
    yielding answers $\mathbf{a} = \{a_1, \ldots, a_n\}$.
\end{enumerate}

The student $S$ is then trained by jointly minimizing a standard
task loss and a distillation loss on the aligned answers:
\begin{equation}\label{eq:student_loss}
    \min\;\mathbb{E}_{(d,\, y_{\text{true}}) \sim \mathcal{D}}
    \Big(\mathcal{L}_{\text{task}}\big(S(d),\, y_{\text{true}}\big)
    + \alpha\, \mathcal{L}_{\text{distill}}(\mathbf{a},\,
    \mathbf{h})\Big)
\end{equation}
where $S(d)$ is the student's prediction,
$\mathcal{L}_{\text{task}}$ is the task-specific loss ({\em e.g.},
cross-entropy), $\mathcal{L}_{\text{distill}}$ measures the
divergence between the answers $\mathbf{a}$ and the student's
last-layer hidden states $\mathbf{h}$, and $\alpha$ controls the
distillation weight. By operating on hidden states rather than
output logits, this formulation avoids output-space
incompatibilities, {\em e.g.}, an LLM producing logits over its
vocabulary while a recommender predicts over candidate items.

\subsection{Detailed Design of the Distillation Module}
\label{sec:distillation_design}

\textbf{Question Augmenter $F_Q$.}
As shown in Fig.~\ref{fig:question_generator_design}, $F_Q$ takes
the student's representation
$\mathbf{x} \in \mathbb{R}^{d_{\text{in}}}$ and produces
teacher-specific questions $\mathbf{q} = \{q_1, \ldots, q_n\}$
along with importance weights $\mathbf{w}$. It operates through
three trainable components, with $d_{\text{emb}}$ and $d_m$ as
hidden dimensions:

\textit{(1) Teacher Embedding
$E \in \mathbb{R}^{n \times d_{\text{emb}}}$}: a matrix of $n$
learnable embeddings, one per teacher. These embeddings encode
each teacher's characteristics and are shared across the question
generation and importance scoring paths.

\textit{(2) Question Projection
$W_M \in \mathbb{R}^{d_{\text{in}} \times d_m \times
d_{\text{emb}}}$}: a tensor that projects $\mathbf{x}$ into $d_m$
components of dimension $d_{\text{emb}}$:
$M = \mathbf{x} W_M \in \mathbb{R}^{d_m \times d_{\text{emb}}}$.
Teacher-conditioned questions are generated via a sequential
dot-product: $EM^\top \in \mathbb{R}^{n \times d_m}$ computes the
alignment between each teacher's embedding and the projected
student representation, and $(EM^\top) M \in \mathbb{R}^{n \times
d_{\text{emb}}}$ produces $n$ question representations, each
reflecting the interaction between the teacher's embedding and the
student's data representation. Each question representation is
then passed through a per-teacher MLP that projects it into
$T_i$'s native input dimension, yielding the final question $q_i$.

\textit{(3) Teacher Importance
$W_F \in \mathbb{R}^{d_{\text{in}} \times d_{\text{emb}}}$}:
computes a dynamic, per-sample importance score for each teacher:
$\mathbf{w} = \sigma(\mathbf{x} W_F \cdot E^\top) \in (0,1)^n$,
where $\sigma$ is the sigmoid function. These scores serve two
purposes: they weight each teacher's contribution in the
distillation loss, and they can optionally be used as computational
filters to skip low-relevance teachers during training
(Sec.~\ref{sec:experiment_result}).

\medskip
\textbf{Answer Augmenter $F_A$.}
As depicted in Fig.~\ref{fig:answer_generator_design}, $F_A$ takes
the set of teacher last-layer hidden states
$\mathbf{p} = \{p_1, \ldots, p_n\}$, where each $p_i$ resides in
$T_i$'s native dimensional space. A per-teacher MLP maps each
$p_i$ into the student's last-layer dimension
$\mathbb{R}^{d_{\text{out}}}$, yielding answer $a_i$. The set of
aligned answers $\mathbf{a} = \{a_1, \ldots, a_n\}$ serves as
soft targets for the distillation loss.

\medskip
\textbf{Distillation Loss.}
The distillation loss in Eq.~\eqref{eq:student_loss} aggregates
the weighted divergence between each teacher's answer and the
student's hidden states:
\begin{equation}\label{eq:distill_loss}
    \mathcal{L}_{\text{distill}}(\mathbf{a}, \mathbf{h})
    = \sum_{i=1}^{n} w_i \, \ell(a_i, \mathbf{h})
\end{equation}
where $\ell$ denotes a per-teacher divergence metric (L2 distance
in our experiments). Both $a_i$ and $\mathbf{h}$ reside in the
student's hidden space ($\mathbb{R}^{d_{\text{out}}}$), so this
formulation is agnostic to the teachers' original output formats.

\subsection{Training the Distillation Module}
\label{sec:training_distillation}

Training the Distillation Module presents a challenge: without
explicit supervision, $F_Q$ and $F_A$ could converge to trivial
representations that minimize the distillation divergence without
capturing task-relevant information. To prevent this, we introduce
a \emph{task regularization} mechanism that anchors the module's
training to the downstream objective.

\begin{figure}[t]
    \centering
    \includegraphics[width=0.6\textwidth]{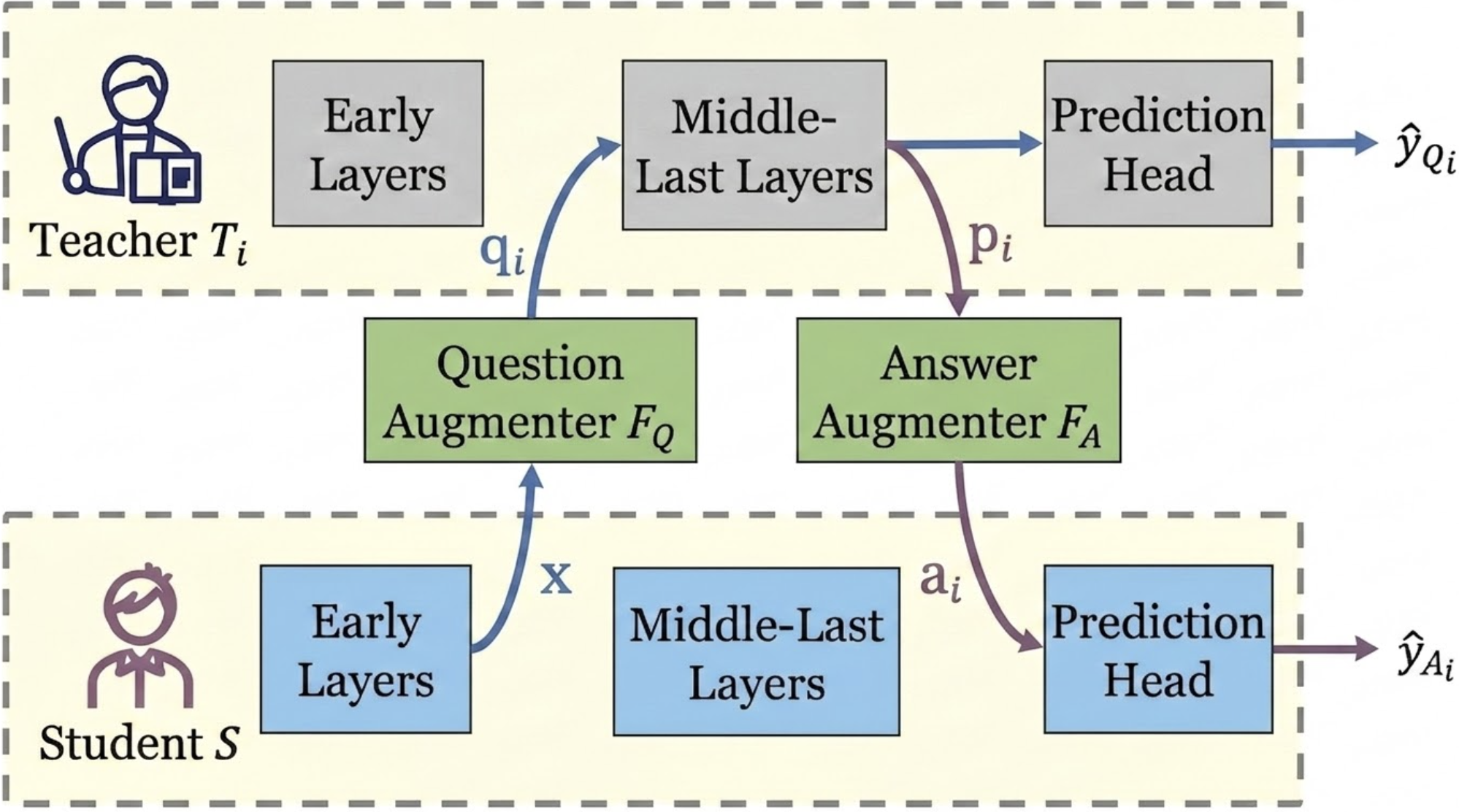}
    \caption{Task Regularizer for task-oriented distillation (shown
for one teacher). Questions $q_i$ from $F_Q$ are
    injected into teacher $T_i$, whose forward pass yields
    prediction $\hat{y}_{Q_i}$. The teacher's hidden states
    $p_i$ are passed through $F_A$ and the student's prediction
    head to yield $\hat{y}_{A_i}$. Both predictions are compared
    against $y_{\text{true}}$. Green blocks are updated by
$\mathcal{L}_{\text{reg}}$; grey blocks are always frozen
(teacher); blue blocks are frozen w.r.t.\
$\mathcal{L}_{\text{reg}}$ but updated by
$\mathcal{L}_{\text{task}}$ (student).}
    \label{fig:train_distill}
\end{figure}

\textbf{Task Regularizer.}
Rather than relying solely on the distillation divergence to guide
$F_Q$ and $F_A$, we explicitly map intermediate representations
back into the prediction space and compare them against
ground-truth labels. The process is illustrated in
Fig.~\ref{fig:train_distill} for a single teacher. For each
teacher $T_i$:
\begin{enumerate}
    \item The question $q_i$ from $F_Q$ is injected into $T_i$ at
    an early processing stage. $T_i$ performs a forward pass,
    producing both its last-layer hidden states $p_i$ and a
    prediction $\hat{y}_{Q_i}$ via its output head.
    \item The hidden states $p_i$ are passed through $F_A$ and then
    through $S$'s prediction head to produce $\hat{y}_{A_i}$.
\end{enumerate}
Both predictions are compared against $y_{\text{true}}$ via the
task loss. By anchoring both the questions (via $\hat{y}_{Q_i}$)
and the answers (via $\hat{y}_{A_i}$) to the label space, this
regularizer forces $F_Q$ to generate discriminative, task-relevant
queries and $F_A$ to extract task-useful information from the
teachers' responses.

The regularization loss over the teacher committee is:
\begin{equation}\label{eq:regularizer}
    \mathcal{L}_{\text{reg}}
    = \sum_{i=1}^{n} w_i \Big(
        \mathcal{L}_{\text{task}}(\hat{y}_{Q_i},\, y_{\text{true}})
      + \mathcal{L}_{\text{task}}(\hat{y}_{A_i},\, y_{\text{true}})
    \Big)
\end{equation}
where $w_i$ denotes the teacher importance weights from $F_Q$,
and $\mathcal{L}_{\text{task}}$ is the same task loss as in
Eq.~\eqref{eq:student_loss}.

\textbf{Necessity of the Regularizer.}
Without the task regularizer, a natural alternative is to update
$F_Q$ and $F_A$ jointly with $S$ via
$\mathcal{L}_{\text{distill}}$. However, this allows the module
to converge to trivial representations that minimize the
distillation divergence without capturing task-relevant
information: a form of shortcut learning. By isolating
$F_Q$ and $F_A$ behind stop-gradients from the student's
loss and supervising them exclusively through
$\mathcal{L}_{\text{reg}}$, we ensure that the generated questions
and answers remain grounded in the downstream objective.
In preliminary experiments, removing the regularizer led to
substantially degraded student performance.

\subsection{End-to-End Training}
\label{sec:end2end}

\begin{algorithm}[t]
\caption{End-to-End Training for \name{}}
\label{alg:training}
\begin{algorithmic}[1]
    \REQUIRE Dataset $\mathcal{D}$, Frozen Teachers $\{T_1,\ldots,T_n\}$
    \ENSURE Trained Student Model $S$
    \STATE Initialize parameters for $S$, $F_Q$, and $F_A$
    \WHILE{not converged}
        \STATE Sample mini-batch $(d, y_{\text{true}}) \sim \mathcal{D}$
        \STATE $\mathcal{L}_{\text{task}} \gets
            \mathcal{L}_{\text{task}}(S(d),\, y_{\text{true}})$
        \STATE $\mathcal{L}_{\text{distill}} \gets
            \sum_{i=1}^{n} w_i\, \ell(a_i,\, \mathbf{h})$
            \hfill \COMMENT{Eq.~\ref{eq:distill_loss}}
        \STATE $\mathcal{L}_{\text{reg}} \gets
            \sum_{i=1}^{n} w_i \big(
            \mathcal{L}_{\text{task}}(\hat{y}_{Q_i},\,
            y_{\text{true}})
            + \mathcal{L}_{\text{task}}(\hat{y}_{A_i},\,
            y_{\text{true}}) \big)$
            \hfill \COMMENT{Eq.~\ref{eq:regularizer}}
        \STATE Update $S$ with
            $\nabla_S(\mathcal{L}_{\text{task}}
            + \alpha\,\mathcal{L}_{\text{distill}})$
        \STATE Update $F_Q, F_A$ with
            $\nabla_{F_Q, F_A}(\beta\,\mathcal{L}_{\text{reg}})$
    \ENDWHILE
    \RETURN $S$
    \COMMENT{$F_Q$, $F_A$, and all teachers are discarded}
\end{algorithmic}
\end{algorithm}

The complete training procedure is summarized in
Algorithm~\ref{alg:training}. The student $S$ and the Distillation
Module ($F_Q$, $F_A$) are optimized jointly but with \emph{separate}
gradient paths. The total loss combines three terms:
\begin{equation}\label{eq:total_loss}
    \mathcal{L}_{\text{total}}
    = \underbrace{\mathcal{L}_{\text{task}}
      + \alpha\,\mathcal{L}_{\text{distill}}}_{\text{updates } S}
    + \underbrace{\beta\,\mathcal{L}_{\text{reg}}}_{\text{updates }
    F_Q,\, F_A}
\end{equation}
where $\mathcal{L}_{\text{task}}$ is the standard task loss
(Eq.~\ref{eq:student_loss}),
$\mathcal{L}_{\text{distill}}$ is the weighted distillation loss
(Eq.~\ref{eq:distill_loss}), $\mathcal{L}_{\text{reg}}$ is the
task regularizer (Eq.~\ref{eq:regularizer}), and $\alpha$, $\beta$
are hyperparameters controlling their relative weights.

\textbf{Gradient Routing.}
As indicated by the underbrace annotations in
Eq.~\eqref{eq:total_loss} and the update rules in
Algorithm~\ref{alg:training} (Lines~7--8), $\mathcal{L}_{\text{task}}$
and $\mathcal{L}_{\text{distill}}$ update only $S$ (with
stop-gradients on $F_Q$ and $F_A$), while $\mathcal{L}_{\text{reg}}$
updates only $F_Q$ and $F_A$ (with stop-gradients on $S$, as marked by the blue blocks in Fig.~\ref{fig:train_distill}). Teachers remain
frozen throughout. This separation isolates the student's task
learning from the Distillation Module's alignment optimization, and
ensures $\mathcal{L}_{\text{reg}}$ is the sole training signal for
$F_Q$ and $F_A$---making the regularizer structurally necessary as
discussed in Sec.~\ref{sec:training_distillation}.

\textbf{Training and Inference Cost.}
The Distillation Module is a training-only scaffold: its parameters
are small relative to the foundation model being distilled ({\em e.g.},
$<$5\% of a 60M-parameter teacher). The dominant training cost is the
teacher forward passes, which are inherent to all distillation
methods; \name{}'s teacher importance weights can further reduce this
cost by skipping low-relevance teachers per sample
(Sec.~\ref{sec:experiment_result}). At inference, the Distillation
Module and all
teachers are discarded---the deployed student is identical in
architecture and latency to an undistilled model.

\section{Experiment}
\label{sec:experiment}

We evaluate \name{} across recommendation and vision tasks to
answer three questions:

\textbf{Q1:} Is distilling directly from foundation models
effective?

\textbf{Q2}: Can standard multi-teacher methods handle diverse
teacher committees?

\textbf{Q3}: Does \name{} enable effective knowledge transfer
from diverse committees where standard methods fail?

\subsection{Experiment Setup}
\label{sec:experiment_setup}
Our teacher committee is composed of one foundation model, and domain expert models that are more similar to the student as complementary teachers. We use TensorFlow and TPU. We summarize model details in Table~\ref{table:mv-teacher-details} and \ref{table:vision-teacher-details}, and explain committee composition and Distillation Module details here. Note Distillation Modules are only used in training but not in inference.

\begin{table}[t]
    \centering
    \begin{minipage}[t]{0.54\textwidth}
        \centering
        \caption{Details on recommendation tasks.}
        \label{table:mv-teacher-details}
        \vspace{1mm} 
        \resizebox{\linewidth}{!}{
            \setlength{\tabcolsep}{3pt}
            \begin{tabular}{c|c|c|c|c}
            \hline
            & Model & Size & Architecture & Input \\
            \hline
            Stu. & MLP-S & 2M & Emb. + MLP [128, 64] & Cat. \\
            \hline
            \multirow{3}{*}{Tea.} & T5 & 76M & Enc-Dec LM & Cat.+Text \\
            \cline{2-5}
            & MLP-M & 3M & Emb. + MLP [512, 256] & Cat. \\
            \cline{2-5}
            & MLP-L & 4M & Emb. + MLP [1024, 512] & Cat. \\
            \hline
            \end{tabular}
        }
    \end{minipage}\hfill
    \begin{minipage}[t]{0.42\textwidth}
        \centering
        \caption{Details on vision tasks.}
        \label{table:vision-teacher-details}
        \vspace{1mm} 
        \resizebox{\linewidth}{!}{
            \setlength{\tabcolsep}{3pt}
            \begin{tabular}{c|c|c|c}
            \hline
            & Model & Size & Input \\
            \hline
            Stu. & ViT-B/16 & 86M & Image \\
            \hline
            \multirow{2}{*}{Tea.} & CLIP & 149M & Image+Text \\
            \cline{2-4}
            & ViT-L & 307M & Image \\
            \hline
            \end{tabular}
        }
    \end{minipage}
    
    \vspace{2mm} 
    \raggedright
    \scriptsize \textit{Note:} Emb.: Embedding, Cat.: Categorical, Enc-Dec LM: Encoder-Decoder Language Model.

\end{table}

\medskip
\noindent\textbf{Recommendation Tasks.} We use MovieLens~\cite{10.1145/2827872} for Rating Prediction. The student model $\mathcal{S}$ follows a standard recommendation architecture~\cite{naumov2019deep}: an MLP with embedding tables. The foundation model is an LLM T5~\cite{raffel2023exploring}, and we follow \cite{kang2023llms} to formalize the input features in natural language and fine-tune T5 on the task dataset. We want domain models to be more similar to $S$, and use larger MLP-based models and train from scratch on the task dataset. Note that models may use different features: MLP-based models use embedding tables and categorical features like user/movie ID, while the LLM takes natural languages and text features like movie titles. To train the Distillation Module, $F_Q$ takes in representations from $S$'s embedding table. Its output questions are consumed by teachers as replacement of (1) the hidden states after embedding table for the MLP teachers (2) the token embedding after the first decoder transformer layer for the T5 teacher. $F_A$ takes in hidden states before the last layer from all the teachers. 

\medskip
\noindent\textbf{Vision Tasks.} We evaluate Image Classification on CIFAR-10, CIFAR-100~\cite{Krizhevsky2009LearningML}, and ImageNet~\cite{russakovsky2015imagenet}. $S$ is a Vision Transformer (ViT)~\cite{dosovitskiy2021image} and we use ViT-Base/16 following~\cite{sariyildiz2024unic}. The foundation model is a pre-trained text-vision model CLIP~\cite{radford2021learning}, and we use it under the zero-shot setting following~\cite{radford2021learning}. The domain model is a larger ViT finetuned on the task datasets. For input features, CLIP uses both the text of the class labels and the images, while ViTs take only images. $F_Q$ takes in the representations of the classification token from the middle layer of $S$. Its output questions are used by teachers as replacement of the classification token in (1) the middle transformer layer for ViTs (2) the transformer image module for CLIP. Inputs for $F_A$ are the representation (1) of the classification token from the last layer for ViTs (2) from the image encoder for CLIP.

\medskip
\noindent\textbf{Baselines.} We compare against methods in two categories:

\noindent\emph{(1) Single-teacher}: Logit Distillation
(\textbf{LD})~\cite{sanh2020distilbert,wu2022tinyvit}, which uses teacher logits as the soft target; Feature Distillation
(\textbf{FD})~\cite{sun2023dimefm}, which uses teacher hidden states with projection layers to bridge dimensional differences; and
Student Customized Knowledge Distillation (\textbf{SCKD})~\cite{9711488}, which examines the alignment between gradients from distillation loss and task loss.

\noindent\emph{(2) Multi-teacher}: Multi-Teacher (MT)~\cite{10.1145/3097983.3098135}, which incorporates multiple teacher networks by averaging soft labels;
Adaptive Multi-Teacher Multi-Level Knowledge Distillation (\textbf{AMTML})~\cite{Liu_2020}, which learns from intermediate layers by multi-group hint; Stochastic Knowledge Distillation to obtain compact BERT style language model (\textbf{AutoSKDBert})~\cite{ding2022skdbert}, which learns a weight for each teacher based on student progress; and Adaptive Ensemble Knowledge Distillation (\textbf{AEKD})~\cite{NEURIPS2020_91c77393}, which dynamically weights teacher models depending on gradient space agreement.
We note that several recent cross-architecture distillation
methods~\cite{li2025fuse,ni2024crossarchi} require
either gradient access to teacher internals, architectural
modifications, or online co-training---assumptions incompatible
with our strictly frozen teacher setting. We restrict comparisons
to methods operating under comparable constraints (frozen teachers,
no architectural modifications) to ensure a fair evaluation.

\medskip
\noindent\textbf{Hyperparameters.}
For \name{}, we grid-search the distillation loss weight $\alpha$
and regularization weight $\beta$ in
Eq.~\eqref{eq:total_loss} over $\{0.1, 0.3, 0.5, 0.8, 1.0\}$,
and the Distillation Module dimensions $d_{\text{emb}}$ and $d_m$
(Sec.~\ref{sec:distillation_design}) over
$d_{\text{emb}} \in \{128, 256, 512\}$ and
$d_m \in \{32, 64, 128\}$.
Standard hyperparameters ({\em e.g.} learning rate, dropout, batch size) are
tuned independently for \name{} and each baseline.

\medskip
\noindent\textbf{Implementation Details.}
For recommendation, we use $d_{\text{emb}} = 128$, $d_m = 64$,
$\alpha = 1.0$, $\beta = 0.5$ for the best committee set-up, and train for 8,000 steps with
batch size 64 using Adagrad (lr = 0.06).
For vision, we use $d_{\text{emb}} = 512$, $d_m = 64$,
$\alpha = 1.0$, $\beta = 1.0$, and fine-tune for 10,000 steps
with SGD (lr = 0.1, cosine decay).
We observe that $\alpha$ and $\beta$ are stable within the same
task and dataset but may vary with committee composition; for
example, $\beta \approx 0.5$ works best when a foundation model
(T5) is present, while $\beta \approx 0.3$ suffices for
homogeneous MLP-only committees. All reported results below use the
best configuration per setting.

\medskip
\noindent\textbf{Reproducibility.}
All evaluations use standard public benchmarks (MovieLens,
CIFAR-10/100, ImageNet). The implementation relies on internal
infrastructure and cannot be publicly released. To support
reproducibility, we provide complete architectural specifications
(tensor dimensions, projection mechanisms, and injection points)
in Sec.~\ref{sec:method} and all hyperparameter configurations
above.

\subsection{Experiment Results}
\label{sec:experiment_result}
\noindent\textbf{Evaluation Metric.}
Directly comparing student performance across tasks is challenging
due to metric incompatibility ({\em e.g.}, lower MSE for rating prediction
vs.\ higher accuracy for classification) and varying task saturation.
To standardize evaluation, we report both the raw task metric and
the \emph{Gap Recovery Rate} (GRR):
\begin{equation}\label{eq:grr}
    \text{GRR} = \frac{S_{\text{base}} - S_{\text{distill}}}
                      {S_{\text{base}} - T_{\text{best}}}
\end{equation}
where $S_{\text{base}}$ is the undistilled student trained from
scratch (serving as the control to isolate the distillation
contribution), $S_{\text{distill}}$ is the distilled student, and
$T_{\text{best}}$ is the best-performing teacher's own prediction.
GRR measures the fraction of the student--teacher gap that is
recovered, accounting for the ``capacity gap" limitations discussed in ~\cite{cho2019efficacy,mirzadeh2020improved}. GRR $>$ 100\% indicates the student surpasses
$T_{\text{best}}$, learning to generalize beyond any single
teacher.

\begin{table*}[t]
    \centering
    \caption{\textbf{Single-Teacher Distillation Results.} Raw metric:
MSE ($\downarrow$) for MovieLens, Accuracy ($\uparrow$) for
vision tasks. GRR defined in Eq.~\ref{eq:grr}; \textbf{bold}
marks the best distilled student per teacher.}
    \label{tab:single_teacher_merged}
    \resizebox{\textwidth}{!}{
    \begin{tabular}{l|cc|cc|cc|cc|cc|cc|cc|cc}
    \toprule
    \textbf{Task} & \multicolumn{4}{c|}{\textbf{MovieLens (MSE $\downarrow$)}} & \multicolumn{4}{c|}{\textbf{CIFAR-10 (Acc $\uparrow$)}} & \multicolumn{4}{c|}{\textbf{CIFAR-100 (Acc $\uparrow$)}} & \multicolumn{4}{c}{\textbf{ImageNet (Acc $\uparrow$)}} \\
    \textit{$S_{base}$} & \multicolumn{4}{c|}{0.8760} & \multicolumn{4}{c|}{0.9771} & \multicolumn{4}{c|}{0.9141} & \multicolumn{4}{c}{0.7720} \\
    \textit{$T_{best}$} & \multicolumn{4}{c|}{0.8210 (MLP\_L)} & \multicolumn{4}{c|}{0.9851 (ViT\_L)} & \multicolumn{4}{c|}{0.9342 (ViT\_L)} & \multicolumn{4}{c}{0.7874 (ViT\_L)} \\ \midrule
    \textbf{Teacher} & \multicolumn{2}{c|}{\textbf{T5}} & \multicolumn{2}{c|}{\textbf{MLP\_L}} & \multicolumn{2}{c|}{\textbf{CLIP}} & \multicolumn{2}{c|}{\textbf{ViT\_L}} & \multicolumn{2}{c|}{\textbf{CLIP}} & \multicolumn{2}{c|}{\textbf{ViT\_L}} & \multicolumn{2}{c|}{\textbf{CLIP}} & \multicolumn{2}{c}{\textbf{ViT\_L}} \\
\midrule
    \textbf{Method} & \textbf{Raw} & \textbf{GRR} & \textbf{Raw} & \textbf{GRR} & \textbf{Raw} & \textbf{GRR} & \textbf{Raw} & \textbf{GRR} & \textbf{Raw} & \textbf{GRR} & \textbf{Raw} & \textbf{GRR} & \textbf{Raw} & \textbf{GRR} & \textbf{Raw} & \textbf{GRR} \\ \midrule
    LD & NA & NA & 0.8461 & 54.5\% & 0.9759 & -15.0\% & 0.9795 & 30.0\% & 0.9223 & 40.8\% & 0.9226 & 42.3\% & 0.7758 & 24.7\% & 0.7752 & 20.8\% \\
    FD & 0.8587 & 31.4\% & 0.8660 & 18.2\% & 0.9787 & 20.0\% & 0.9797 & 32.5\% & 0.9226 & 42.0\% & 0.9210 & 34.3\% & 0.7754 & 22.1\% & 0.7752 & 20.8\% \\
    SCKD & 0.8605 & 28.2\% & 0.8464 & 53.8\% & 0.9794 & 28.8\% & 0.9796 & 31.3\% & 0.9156 & 7.5\% & 0.9162 & 10.4\% & 0.7760 & 25.3\% & 0.7755 & 22.7\% \\ \midrule
    \name & 0.8546 & 38.9\% & \textbf{0.8358} & \textbf{73.1}\% & \textbf{0.9810} & \textbf{37.5}\% & 0.9800 & 36.3\% & \textbf{0.9241} & \textbf{49.8}\% & 0.9227 & 42.8\% & 0.7840 & 79.9\% & \textbf{0.7845} & \textbf{81.2}\% \\ \bottomrule
    \end{tabular}
    }
\end{table*}

\medskip
\noindent\textbf{Q1: Foundation models can be poor direct teachers.}
In Table~\ref{tab:single_teacher_merged}, we summarize the teaching performance on a single-teacher basis. We first observe that
foundation models (T5, CLIP) can often underperform as direct
distillation teachers. T5 achieves at most 38.9\% GRR on MovieLens (with \name{}),
and CLIP even yields \emph{negative} GRR on CIFAR-10 ($-$15.0\% with
LD), meaning the student is \emph{worse} after distillation. This
is unsurprising: the large capacity gap and distribution mismatch
between foundation models and domain-specific compact students make standard
distillation ineffective. In contrast, domain-specific teachers
(MLP-L, ViT-L) are often more effective ({\em e.g.}, MLP-L: 73.1\% GRR on
MovieLens), but remain limited by their narrower knowledge scope. 
We also see that even in the single-teacher setting, \name{} consistently
outperforms all baselines regardless of teacher type, confirming
that the interactive QA mechanism benefits distillation even
without a committee.

\begin{table*}[t]
\centering
\caption{\textbf{Multi-teacher distillation results.} Each task uses a
diverse 2-teacher committee with 1 foundation model and 1 domain model. GRR values $>$100\% indicate the
student surpasses the best single teacher $T_{\text{best}}$. \textbf{Bold} denotes the best distilled student per committee.}

\label{tab:multi_teacher_merged}
\resizebox{0.9\textwidth}{!}{
\begin{tabular}{l|cc|cc|cc|cc}
\toprule
\textbf{Task} 
    & \multicolumn{2}{c|}{\textbf{MovieLens (MSE $\downarrow$)}} 
    & \multicolumn{2}{c|}{\textbf{CIFAR-10 (Acc $\uparrow$)}} 
    & \multicolumn{2}{c|}{\textbf{CIFAR-100 (Acc $\uparrow$)}} 
    & \multicolumn{2}{c}{\textbf{ImageNet (Acc $\uparrow$)}} \\ 
$S_{\text{base}}$ 
   & \multicolumn{2}{c|}{0.8760} 
   & \multicolumn{2}{c|}{0.9771} 
   & \multicolumn{2}{c|}{0.9141} 
   & \multicolumn{2}{c}{0.7720} \\
$T_{best}$
   & \multicolumn{2}{c|}{0.8210 (MLP\_L)} 
   & \multicolumn{2}{c|}{0.9851 (ViT\_L)} 
   & \multicolumn{2}{c|}{0.9342 (ViT\_L)} 
   & \multicolumn{2}{c}{0.7874 (ViT\_L)} \\ 
\midrule
\textbf{Committee} 
   & \multicolumn{2}{c|}{\textbf{MLP\_L + T5}} 
   & \multicolumn{2}{c|}{\textbf{CLIP + ViT\_L}} 
   & \multicolumn{2}{c|}{\textbf{CLIP + ViT\_L}} 
   & \multicolumn{2}{c}{\textbf{CLIP + ViT\_L}} \\ 
\midrule
    
\textbf{Method} & \textbf{Raw} & \textbf{GRR} & \textbf{Raw} & \textbf{GRR} & \textbf{Raw} & \textbf{GRR} & \textbf{Raw} & \textbf{GRR} \\ 
\midrule
    MT & 0.8429 & 60.2\% & 0.9761 & -12.5\% & 0.9228 & 43.3\% & 0.7753 & 21.4\% \\
    AMTML & 0.8591 & 30.7\% & 0.9782 & 13.7\% & 0.9200 & 29.4\% & 0.7754 & 22.1\% \\
    AutoSKDBert & 0.8411 & 63.5\% & 0.9763 & -10.0\% & 0.9237 & 47.8\% & 0.7840 & 77.9\% \\
    AEKD & 0.8522 & 43.3\% & 0.9796 & 31.3\% & 0.9172 & 15.4\% & 0.7758 & 24.7\% \\ \midrule
    \name & \textbf{0.8132} & \textbf{114.2}\% & \textbf{0.9837} & \textbf{82.5}\% & \textbf{0.9288} & \textbf{73.1}\% & \textbf{0.7870} & \textbf{90.9}\% \\ 
\bottomrule
    \end{tabular}
    }
\end{table*}
\medskip
\noindent\textbf{Q2: Naive multi-teacher methods fail with
diverse committees.}
Table~\ref{tab:multi_teacher_merged} reports the 2-teacher committee performance with 1 foundation model and 1 domain teacher model. We see that standard multi-teacher methods struggle with such diverse committee. For example, simple soft-label averaging (MT) degrades the student
to $-$12.5\% GRR on CIFAR-10---\emph{worse than no distillation
at all}. On MovieLens, MT with the diverse committee (T5 + MLP-L)
achieves only 60.2\% GRR, substantially below MLP-L alone
(73.1\%), demonstrating that naively combining a foundation model
with a domain expert \emph{hurts} more than it helps. More
sophisticated baselines (AMTML, AEKD, AutoSKDBert) may partially
mitigate this issue but often exhibit high performance variance across
tasks, underscoring the difficulty of accommodating heterogeneous
teacher expertise with fixed aggregation strategies.

\medskip
\noindent\textbf{Q3: \name{} makes diversity work.}
\name{} achieves consistent and substantial improvements over all
multi-teacher baselines across every dataset from Table~\ref{tab:multi_teacher_merged}. On MovieLens, the
diverse committee (T5 + MLP-L) with \name{} achieves
\textbf{114.2\% GRR}---the student \emph{surpasses} the best
single teacher, recovering more than the full performance gap.
This is a dramatic improvement over the same committee with other baselines
(highest 63.5\%) and even over the best single-teacher result (73.1\%).
Across vision tasks, \name{} achieves 73.1--90.9\% GRR,
outperforming all baselines and single-teacher results by large margins. These results
confirm that the interactive, teacher-specific communication
mechanism of \name{} successfully bridges the challenges of
heterogeneous teacher distillation: despite a $38\times$
compression ratio in recommendation (76M teacher to 2M student)
and $3.6\times$ in vision (307M to 86M), the distilled student
matches or exceeds its foundation teachers' performance.

\subsection{Ablation and Analysis}
\label{sec:exp_ablation}

We investigate two design aspects of \name{}: the impact of
committee composition on distillation quality, and the use of
teacher importance scores for training efficiency. We use
MovieLens as a representative case study.

\begin{table}[t]
\caption{Effect of committee composition on MovieLens
(MSE $\downarrow$). \textbf{Bold} shows GRR values $>$100\% indicating the
student surpasses the best single teacher $T_{\text{best}}$.}
\label{tab:committee_ablation}
\centering
\begin{tabular}{llccc}
\toprule
\textbf{Committee} & \textbf{Type}
  & \textbf{Teacher Count} & \textbf{MSE} & \textbf{GRR} \\
\midrule
T5              & Foundation  & 1 & 0.8546 & 38.9\% \\
MLP\_M          & Domain      & 1 & 0.8365 & 71.8\% \\
MLP\_L          & Domain      & 1 & 0.8358 & 73.1\% \\
\midrule
MLP\_L + MLP\_M & Homogeneous & 2 & 0.8247 & 93.3\% \\
MLP\_M + T5     & Diverse     & 2 & 0.8197 & \textbf{102.4\%} \\
MLP\_L + T5     & Diverse     & 2 & 0.8132
  & \textbf{114.2\%} \\
\midrule
MLP\_L + MLP\_M + T5
  & Diverse & 3 & 0.8146 & \textbf{111.6\%} \\
\bottomrule
\end{tabular}
\end{table}

\medskip
\noindent\textbf{Committee Composition.}
Table~\ref{tab:committee_ablation} reveals two findings.
First, committee \emph{diversity} is a stronger driver of
effective distillation than committee \emph{size}: the
diverse 2-teacher committee (MLP\_L + T5, 114.2\% GRR)
substantially outperforms the homogeneous committee
(MLP\_L + MLP\_M, 93.3\% GRR), despite both having two
teachers. This supports our hypothesis that combining a
foundation model with a domain-specific expert provides
complementary knowledge that any single paradigm cannot.
Second, every diverse committee---regardless of size---enables
the student to \emph{surpass} the best single teacher
($>$100\% GRR), while homogeneous and single-teacher
settings do not. Adding a third teacher
(MLP\_L + MLP\_M + T5, 111.6\%) maintains this advantage
but provides no further gain over the best two-teacher
pair, likely because MLP\_M's domain expertise largely
overlaps with MLP\_L, contributing redundant knowledge.

\begin{figure}[t]
    \centering
    \subfigure[Mean Utilization]{
    \includegraphics[width=0.3\textwidth]{image/experiment/bar_w.pdf}
    }
    \subfigure[Normalized Batch Utilization]{
    \includegraphics[width=0.55\textwidth]{image/experiment/line_w.pdf}
    }
    \caption{Teacher Utilization. Analysis on a randomly sampled MovieLens batch. If teacher importance $w<0.6$, set $w=0$ as the teacher is skipped for that sample. (a): Mean $w$ across the batch (b): Normalized $w$ (L1-norm) over data samples, where the heavier T5 is skipped more often than MLPs to save computation.}
    \label{fig:importance_score}

\end{figure}
\medskip
\noindent\textbf{Teacher Selection via Importance Scores.}
A potential concern with larger committees is increased
training cost from additional teacher forward passes.
The per-sample importance scores $w_i \in (0, 1)$ from $F_Q$
(Sec.~\ref{sec:distillation_design}) provide a natural
remedy: by applying a threshold, the student only queries
``relevant'' teachers for each sample, skipping the forward
pass for the rest. For the 3-teacher MovieLens committee
(MLP\_L + MLP\_M + T5), a threshold of 0.6 saves $\sim$30\%
of teacher forward passes with no loss in performance, as T5
is skipped more frequently than the MLP teachers.
Fig.~\ref{fig:importance_score} visualizes the per-teacher
utilization probabilities under threshold filtering over a
randomly sampled data batch.

\section{Conclusion and Discussion}

 We presented \name{}, an interactive distillation framework that
bridges the capacity and modality gaps between foundation models
and compact deployment-ready students. By constructing a diverse
teacher committee and introducing a learnable Distillation Module,
\name{} achieves semantic alignment across heterogeneous
architectures without requiring architectural modifications or
co-training of the teachers. Our experiments show that \name{}
consistently recovers—and in several cases exceeds—the full
teacher--student performance gap (GRR $>$ 100\%), compressing
models by up to $38\times$ while preserving identical inference
efficiency. These results suggest that teacher-specific,
interactive knowledge extraction is essential for effective
multi-source distillation, particularly when teacher expertise
spans different modalities and architectures.

We acknowledge several limitations of the current work.
First, we report descriptive metrics without formal statistical
significance testing ({\em e.g.}, paired t-tests), which is important
given the narrow absolute margins on saturated benchmarks.
Second, we evaluate a single student architecture per domain;
systematically varying student capacity would clarify how
\name{}'s benefits scale with the teacher--student gap.
Third, our committees comprise 2--3 teachers; investigating
larger committees and diminishing marginal returns is an
important direction. Looking ahead, \name{}'s frozen-teacher
design makes it naturally suited for distillation from
API-only foundation models, an increasingly common deployment
scenario that we plan to explore in future work.



\bibliographystyle{splncs04}
\bibliography{example_paper}

\end{document}